\documentclass{article}
\pdfoutput=1
    \usepackage[preprint, nonatbib]{neurips_2020}

\usepackage[utf8]{inputenc} % allow utf-8 input
\usepackage[T1]{fontenc}    % use 8-bit T1 fonts
\usepackage{url}            % simple URL typesetting
\usepackage{booktabs}       % professional-quality tables
\usepackage{amsfonts}       % blackboard math symbols
\usepackage{nicefrac}       % compact symbols for 1/2, etc.
\usepackage{microtype}      % microtypography
\usepackage{graphicx}
\usepackage{bmpsize}
\usepackage{subcaption}
\usepackage{float}

\usepackage{amsmath}

\title{Temporal Autoencoder with U-Net Style Skip-Connections for Frame Prediction}

\author{Jay Santokhi\hspace{5mm}Pankaj Daga\hspace{5mm}Joned Sarwar\hspace{5mm}Anna Jordan\hspace{5mm}Emil Hewage\\[1mm]
      Alchera Data Technologies Ltd \\
      Cambridge, CB2 1NN\\
      \texttt{\{jay,pankaj,joned,anna,emil\}@alcheratechnologies.com} 
}

\begin{document}

\maketitle

\begin{abstract}
Finding sustainable and novel solutions to predict city-wide mobility behaviour is an ever-growing problem given increased urban complexity and growing populations. This paper seeks to address this by describing a traffic frame prediction approach that uses Convolutional LSTMs to create a Temporal Autoencoder with U-Net style skip-connections that marry together recurrent and traditional computer vision techniques to capture spatio-temporal dependencies at different scales without losing topological details of a given city. Utilisation of Cyclical Learning Rates is also presented, improving training efficiency by achieving lower loss scores in fewer epochs than standard approaches.
\end{abstract}
\section{Introduction}
The traffic prediction problem presented by Traffic4Cast 2020 is at first glace a time series problem. Instinctively one would assume solutions using a form of Recurrent Neural Network (RNN) to help capture temporal features. However, the presentation of time series probe data as image frames by mapping onto geographical locations introduces a spatial aspect to this problem. Approaching this as Frame Prediction makes the problem both simultaneously image-to-image translation \cite{Isola2017} and sequence-to-sequence \cite{SutskeverGoogle2014}.

Solutions for Traffic4Cast 2019 seemed to fall into one of two camps; the first opting strictly for an image-to-image approach using predominately U-Net \cite{Ronneberger2015} based designs \cite{Choi, Martin2019, Liu2019} and the others aiming to incorporate both recurrent and computer vision techniques to capture spatio-temporal features \cite{Herruzo2019, Nguyen2019, Yu2019}. 

Given these findings and the changes in data presentation from the 2019 challenge to the 2020 challenge, this paper aims to showcase a Temporal Autoencoder to make predictions on traffic speed and volume up to one hour into the future for the target cities. This is achieved by using Convolutional LSTM layers to capture spatio-temporal patterns while also utilising U-Net style skip-connections to preserve topological details across the network.

\section{Data and Problem Definition}
The competition dataset \cite{KoppMichaelKopp2020} consists of training, validation and test sets for each of the 3 challenge cities: Berlin, Istanbul and Moscow. The  area of each city is represented as a 495x436 grid where each pixel encompasses a region of 100 metres x 100 metres. 
% The static data contains road map attributes and points of interest given as a single tensor of shape (495, 436, 7). The first 2 channels are a representation of the junction count while the next 5 encode the number of eating, drinking and entertainment places, the number of hospitals, the number of parking places, the number of shops and the number of public transport venues in that order.

The dataset is made up of temporal probe data of speed and volume at different headings, grouped into 5 minutes time bins.
% collected from GPS positions reported by a large fleet of probe vehicles and from live incidents feeds. 
For each city, the training set consists of 181 training days starting on January 1st 2019 through to June 30th 2019. The validation set contains 18 days sampled from the remaining months of 2019 (July - December). Each day is represented as a (288, 495, 436, 9) tensor. The first dimension gives the number of 5 minute time bins in a 24 hour period. The second and third dimension represent the grid area of the city. The first two of the 9 channels encode aggregated volume and average speed of all underlying probes whose heading is North East, the next two provide the same aggregation of volume and speed heading North West, the following two for South East and South West, respectively. The final channel encodes road incidents. For each of the channels encoding volume, 0 refers to zero volume and values 1-255 represent a linear interpolation of real world probe points up to a maximum. For each of the channels encoding average speed, 0 refers to either no probes or speed lower than a minimum and values 1-255 linearly interpolate the interval [v, V] where v and V denote minimum and maximum speed respectively.

% Testing set tensors are of sizes (m, 12, 495, 436, 9) where m denotes the number of predictions required for that given day and the 12 indicates twelve successive frames with time bin intervals of 5 minutes, totalling 1 hour of data. Predictions must be of size (m, 6, 495, 436, 8) reflecting the 6 time bins of interest: 5, 10, 15, 30, 45 and 60 minutes of the 8 dynamic data channels; predictions for 9th channel (road incidents) are not required. 
In total 500 predictions spread over 163 days are required with mean squared error (MSE) being used as the scoring metric. 

\section{Method}
% This section will provide various details on the methodology, discussing data preparation, using Convolutional LSTMs, final model architecture, training process and finally the prediction process.
\subsection{Data Preparation}
Data preparation for a time-series image sequences involves two main processes: sampling strategy and pre-processing.

For sampling strategy, two main approaches were explored: non-overlapping and overlapping. The testing set provides up to 1 hour of data for use in making predictions for the upcoming hour, thus an input sequence length of 12 was chosen to make full use of the data available. An output sequence length of 12 was also chosen.

Using equation (1), the non-overlapping strategy, with input and output sequence lengths ($S_i$ and $S_o$) of 12, yields 12 training sequences per day, $S_{day}$. Thus $181\times12=2172$ training sequences per city.
% Using equation (1) the non-overlapping strategy, with and input sequence, $S_i$ of 12 and an output sequence, $S_o$ of 12 yields 12 training pairs per day, $S_{day}$ and thus $181 \times 12 = 2172$ total training sequences per city (12 frames for training and 12 for testing).
\begin{equation}
    S_{day} = \frac{288}{S_i + S_o}
\end{equation}
The overlapping strategy, equation (2), makes use of every possible sequence by setting the sliding window to 1. This yields a total of 47965 training sequences per city. 
\begin{equation}
    S_{day} = 288 - (S_i + S_o - 1) 
\end{equation}
The overlapping approach was trialled initially with varying sliding window sizes but given significantly increased training times for minimal improvements it was decided to use non-overlapping to aid the overall training speed.

For pre-processing the data was normalised to between 0 and 1 and the 9th channel specifying road incidents was not used, however in future work these attributes could be used as a form of latent embedding.

% 181 x [(288 - (12+12))/2*overlapping)+1]
\subsection{Capturing Spatio-Temporal Features}
Convolutional LSTMs \cite{Shi2015} are a variant of LSTMs (Long Short-Term Memory) that replaces matrix multiplication with convolution operations at each gate in the LSTM cell. This can be seen when comparing equation (\ref{eq:lstm}) (LSTM) and equation (\ref{eq:convlstm}) (ConvLSTM). Both are a special kind of RNN capable of learning long-term dependencies, however the main difference is input dimension; standard LSTMs takes 1D data and does not support spatial sequence data such as frames of a video, whereas Convolutional LSTMs take 3D data, capturing underlying spatial features through convolution operations as well as temporal features from its LSTM design. This ability to capture spatio-temporal features make it ideal to form the basis of a frame prediction model.
%  By extending the fully connected
% LSTM (FC-LSTM) to have convolutional structures in both the input-to-state and
% state-to-state transitions, we propose the convolutional LSTM (ConvLSTM) and
% use it to build an end-to-end trainable model for the precipitation nowcasting problem. Experiments show that our ConvLSTM network captures spatiotemporal
% correlations better and consistently outperforms FC-LSTM and the state-of-the
\begin{equation}\label{eq:lstm}
\begin{gathered}
  i_{t} = \sigma(W_{i}x_{t} + U_{i}h_{t-1} + V_{i} \circ c_{t-1} + b_{i}) 
  \\
  f_{t} = \sigma(W_{f}x_{t} + U_{f}h_{t-1} + V_{f} \circ c_{t-1} + b_{f}) 
  \\
  c_{t} = f_{t} \circ c_{t-1} + i_{t} \circ tanh(W_{c}x_{t} + U_{c}h_{t-1} + b_{c})
  \\
  o_{t} = \sigma(W_{o}x_{t} + U_{o}h_{t-1} + V_{o} \circ c_{t} + b_{o}) 
  \\
  h_{t} = o_{t} \circ tanh(c_{t})
%   \\
%   \text{where~$R$ is Racoon,~$E$ is Elephant}
\end{gathered}
\end{equation}

\begin{equation}\label{eq:convlstm}
\begin{gathered}
  i_{t} = \sigma(W_{i}*\mathcal{X}_{t} + U_{i}*\mathcal{H}_{t-1} + V_{i} \circ \mathcal{C}_{t-1} + b_{i}) 
  \\
  f_{t} = \sigma(W_{f}*\mathcal{X}_{t} + U_{f}*\mathcal{H}_{t-1} + V_{f} \circ \mathcal{C}_{t-1} + b_{f}) 
  \\
  \mathcal{C}_{t} = f_{t} \circ \mathcal{C}_{t-1} + i_{t} \circ tanh(W_{c}*\mathcal{X}_{t} + U_{c}*\mathcal{H}_{t-1} + b_{c})
  \\
  o_{t} = \sigma(W_{o}*\mathcal{X}_{t} + U_{o}*\mathcal{H}_{t-1} + V_{o} \circ \mathcal{C}_{t} + b_{o}) 
  \\
  \mathcal{H}_{t} = o_{t} \circ tanh(\mathcal{C}_{t})
\end{gathered}
\end{equation}

\begin{table}[H]
  \caption{LSTM Symbol Meanings}
  \label{symbols}
  \centering
  \begin{tabular}{lll}
    \toprule
    \cmidrule(r){1-2}
    Symbol       & Meaning \\
    \midrule
    $i_t$     & input gate, weight of acquiring new information  \\ 
    $f_{t}$   & forget gate, weight of remembering old information  \\
    $c_t$     & cell state  \\
    $o_t$     & output gate  \\ 
    $h_t$   & hidden state  \\
    $W_g, V_g, U_g, b_g$   & weight matrices and bias vector for a gate, $g$\\
    $x_t, \mathcal{X}_t$     & input vector, input matrix (image)  \\
    $h_t, \mathcal{H}_t$   & hidden state vector, hidden state matrix (image)  \\
    $c_t, \mathcal{C}_t$     & cell state vector, cell state matrix (image)\\
    $\sigma$     & sigmoid function  \\ 
    $\circ$    & hadamard product  \\
    $*$ & convolution operation \\
    $tanh$ & hyperbolic tangent \\
    \bottomrule
  \end{tabular}
\end{table}

\subsection{Model Architecture}
The final model architecture is at its core an Autoencoder. In order to capture both spatial and temporal features, Convolutional LSTM layers were used to build up the Autoencoder skeleton with 3 layers acting as the encoder and 3 acting as the decoder. This skeleton will be referred to as the `base model'.

\begin{figure}[ht]
  \centering
%   \fbox{\rule[-.5cm]{0cm}{4cm} \rule[-.5cm]{4cm}{0cm}}
  \includegraphics[width=0.875\linewidth]{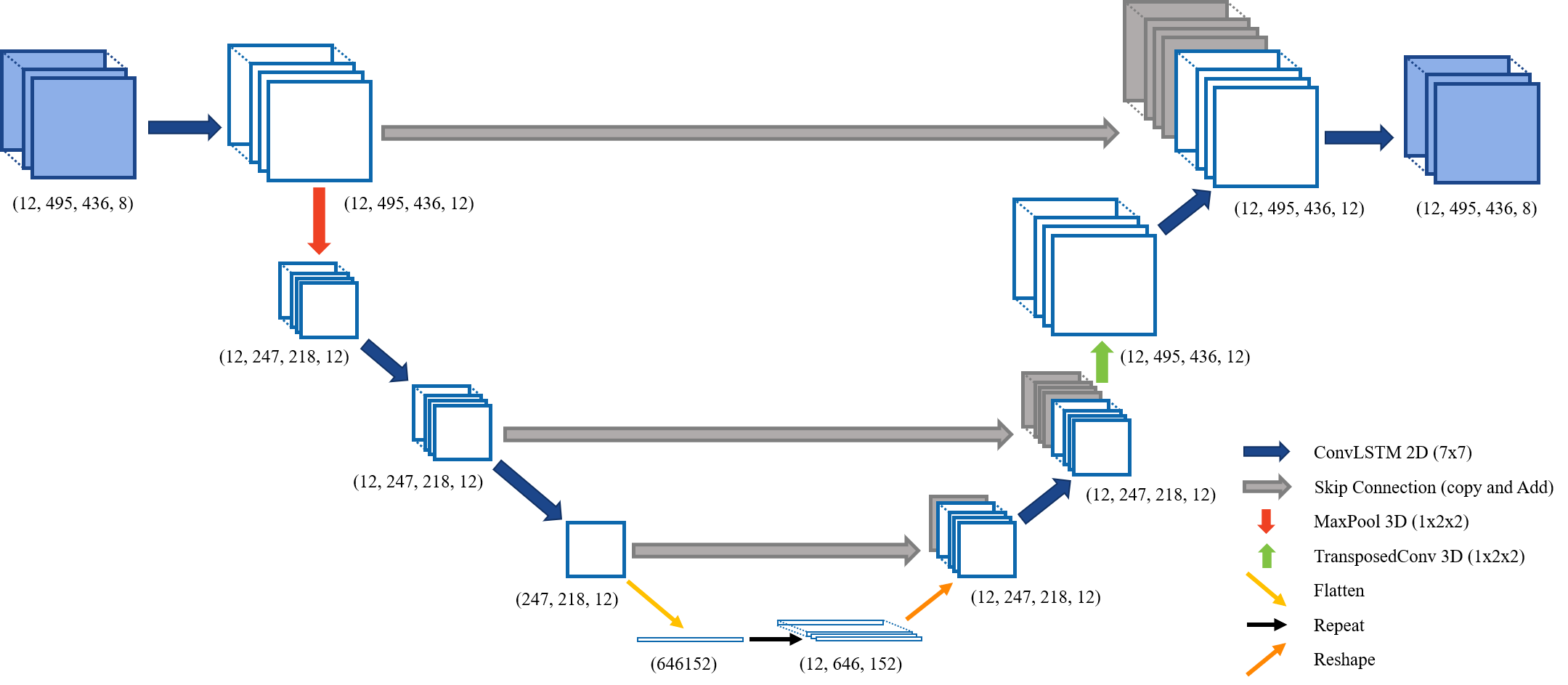}
  \caption{Final Model Architecture (base model with down-sampling and skip-connections)}
  \label{fig:model}
\end{figure}

The overall structure of this base model was iterated on to achieve the final model, seen in figure \ref{fig:model}. Initially a standard 3D convolutional layer was added prior to the first Convolutional LSTM layer to act as a form of feature sampling, however this model did not reach a training loss similar to the base model even with a considerable number of additional epochs.

Down-sampling via 3D Max-Pooling was used to ease the training workload especially given the compute resources available. For up-sampling, Transposed Convolutions were used over standard interpolation up-sampling methods to enable the network to learn up-sampling optimally via learnable parameters. This improved training efficiency and reached similar training and validation losses to the base model in fewer epochs but was still inferior in overall training and validation loss. It is felt that this may be due to loss of finer details in city topology when down-sampling and up-sampling.

To improve on this potential loss in topology details skip-connections were added, used in a similar way to how they are presented in U-Net \cite{Ronneberger2015}. By utilising skip-connections in this way, topology information would not be lost when both down-sampling to a latent space and later reconstructing using Transposed Convolutions.

This iteration achieved the lowest training loss and was further improved when trained using cylical learning rates. The final model architecture can be seen in figure \ref{fig:model} with the tensor shapes at each stage also given.
% \vspace{-0.5cm}
\subsection{Training}
% The model described was trained using 4 NVIDIA 1080 Titans with a batch size of 4 with each GPU taking a single batch using the Tensorflow Mirrored Strategy for GPU pararellisation. This was trained for 28 epochs taking 9.5 hours.
The model described was trained using 4 NVIDIA 1080 Titans with a batch size of 4
for 28 epochs taking a total of 9.5 hours. Due to these long training times, exploring parameter space for an optimal learning rate proved to be unfeasible. As a consequence of this it was opted to make use of a cyclical learning rate approach \cite{Smith2017}.

A cyclical learning rate oscillates between a minimum and maximum learning rate in one of three ways: a standard triangular oscillation where the learning rate increases and decreases linearly at the same rate between an upper and lower bound, a triangular method with amplitude dampening that scales the initial amplitude by half each oscillation and a the third method that scales the initial amplitude exponentially (given by a specified parameter) at each oscillation. One needs to be careful when selecting the upper bound for the learning rate; a maximum learning rate that is too high can cause weights to `blow' up leading to training and validation losses increasing. This approach to learning rates led to improved training efficiency, achieving a smaller training loss in fewer epochs than the traditional approach (of a small learning rate over a large number of epochs) that was used initially.
 
\begin{figure}[ht]
  \centering
  \begin{subfigure}[b]{0.435\linewidth}
    \includegraphics[width=\linewidth]{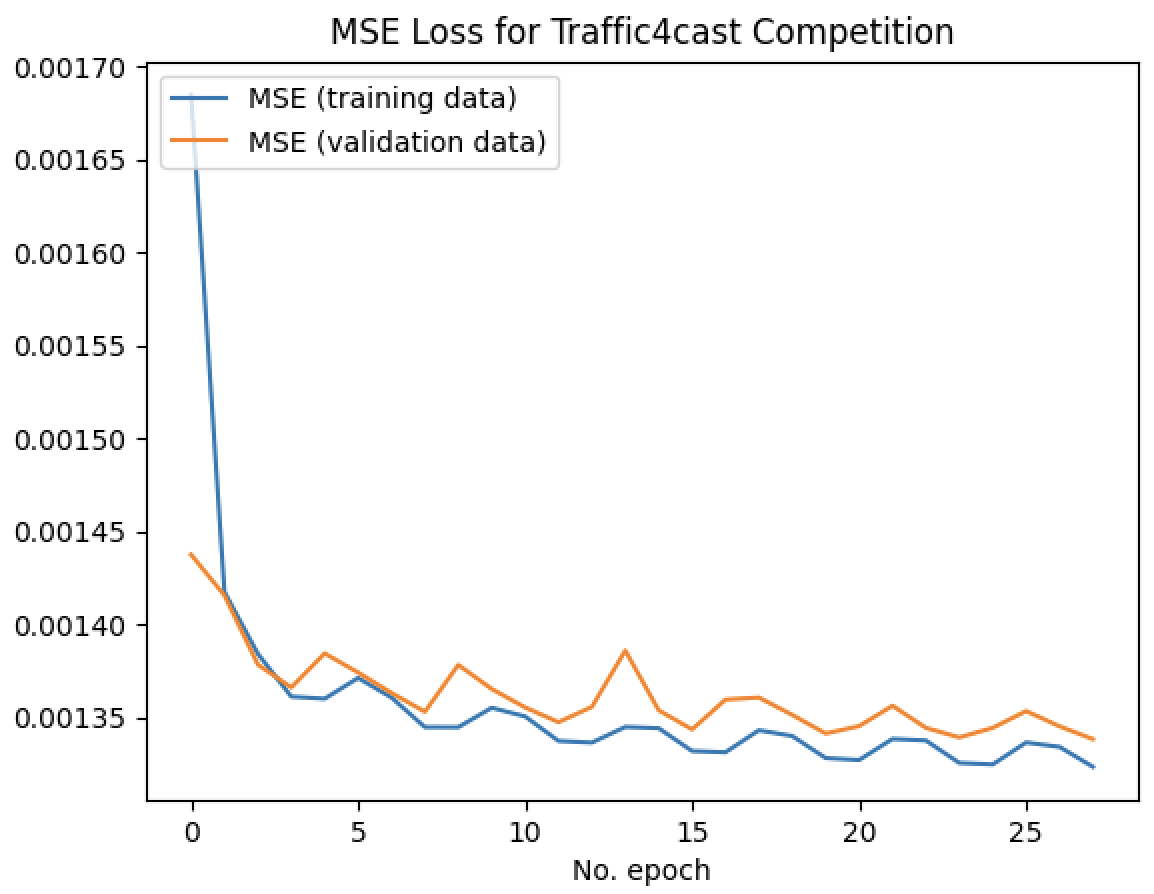}
    \caption{Train and Val Loss for Berlin}
  \end{subfigure}
  \begin{subfigure}[b]{0.435\linewidth}
    \includegraphics[width=\linewidth]{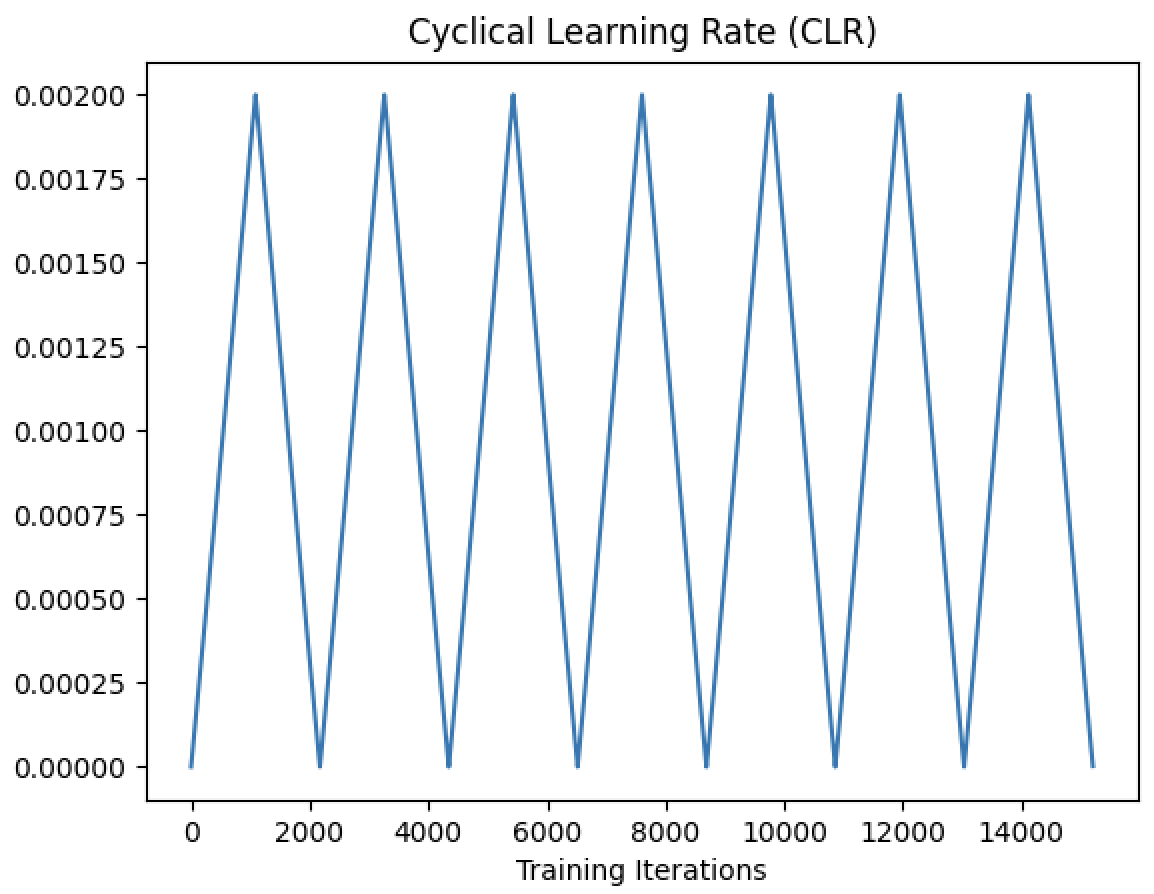}
    \caption{Triangular cyclical learning rate}
  \end{subfigure}
  \caption{Training and Validation Loss with Triangular Cyclical Learning rate}
  \label{fig:training}
\end{figure}

The standard triangular method was adopted for this task making seven full oscillations during the 28 epochs. A maximum learning rate of 0.002 was chosen with a minimum of $1\times10^{-7}$ using Adam as the learning optimiser. Figure \ref{fig:training}a and b shows the training and validation losses during this training process as well as the nature of the learning rate oscillations.

Various literature suggest using the Adagrad learning optimiser for training models on sparse data \cite{Ruder2016}, however for this particular task the loss when using Adagrad would not converge as fast as Adam even when using cyclical learning rates. 
Manageable training times was a major factor to allow regular model iteration and thus using Adam with a cyclical learning rate was the optimal approach. 

Each city was trained separately on the same model architecture resulting in 3 models. Given significant topology differences in each of the cities and the sheer fact that naturally driving behaviour and driving culture is considerably different across the 3 cities (each in different countries), it did not seem wise to encompass these factors into a single `all-in-one' model. For this reason 3 separate models were trained, keeping model architecture, training procedure and hyperparameters exactly the same, from minimum and maximum learning rate to batch size and number of epochs.
\clearpage
\subsection{Predictions}
Predictions up to one hour in the future were made with 6 time bins being of interest, those being: 5, 10, 15, 30, 45 and 60 minutes. Given that each time bin represents 5 minutes, 12 time bin predictions were calculated. Time bins representing 20, 25, 35, 40, 50 and 55 minutes were not required for the task however, for use in a real world setting it is felt that all 12 time bin predictions for the hour would be used. Given that only 6 were of interest for the task they were selected from the 12 prediction time bins rather than forcing the model to learn only those specific bins.

A binary mask was used to smooth the results by removing predictions that may appear where no road network exists, resulting in a minor reduction in MSE loss. It was thought best not to incorporate the binary mask into the training procedure to ensure that the model would learn the topology of the given road networks without the safety net of a binary mask. 

\begin{figure}[H]
  \centering
  \begin{subfigure}[b]{0.49\linewidth}
    \includegraphics[trim=0.5cm 0.5cm 1cm 1cm, clip=true,width=\linewidth]{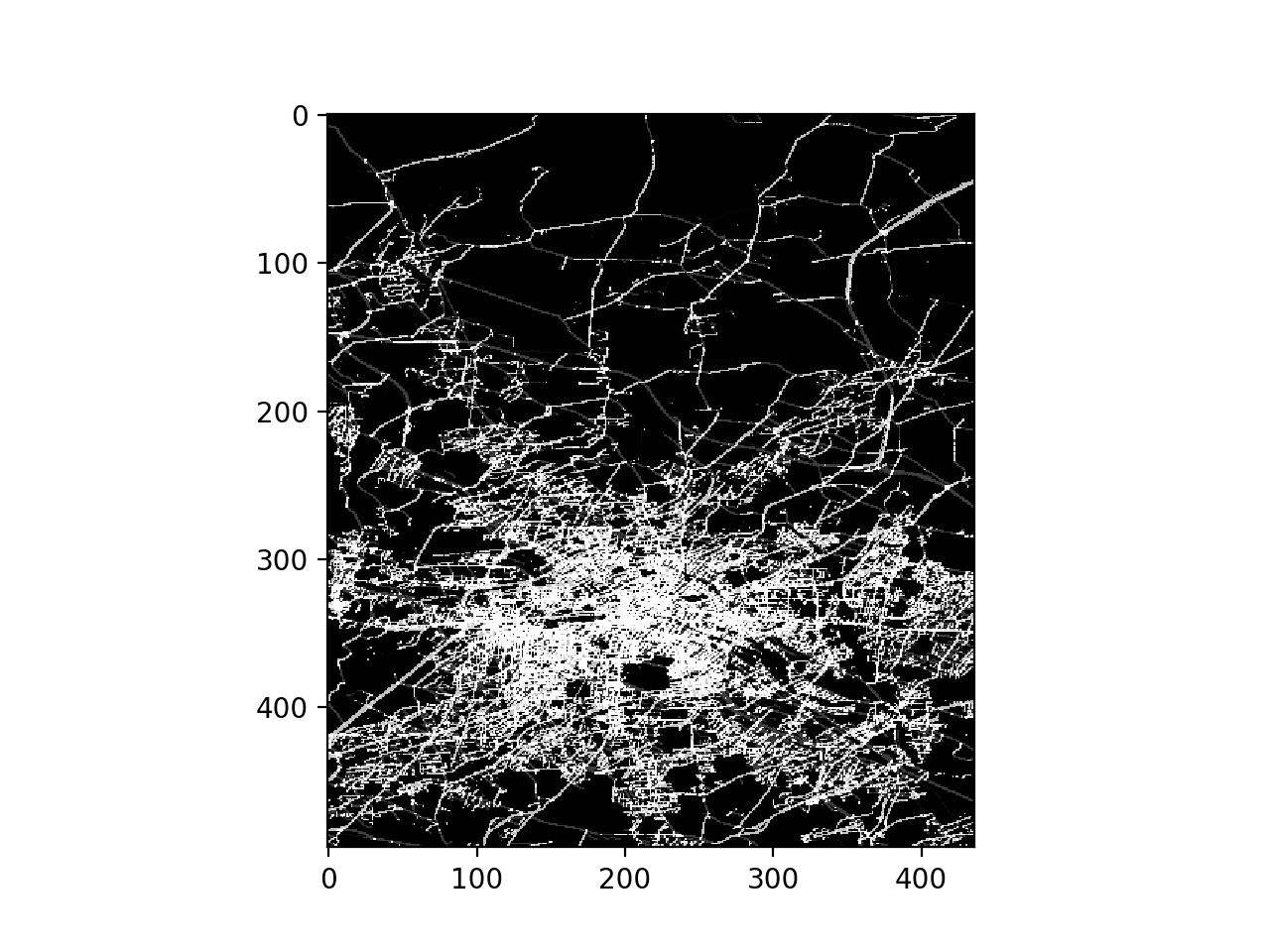}
    % \includegraphics[width=\linewidth]{figures/Screenshot 2020-11-05 at 18.21.02.png}
    % \vspace{-5mm}
    \caption{1 day of data}
  \end{subfigure}
  \begin{subfigure}[b]{0.49\linewidth}
    \includegraphics[trim=0.5cm 0.5cm 1cm 1cm, clip=true,width=\linewidth]{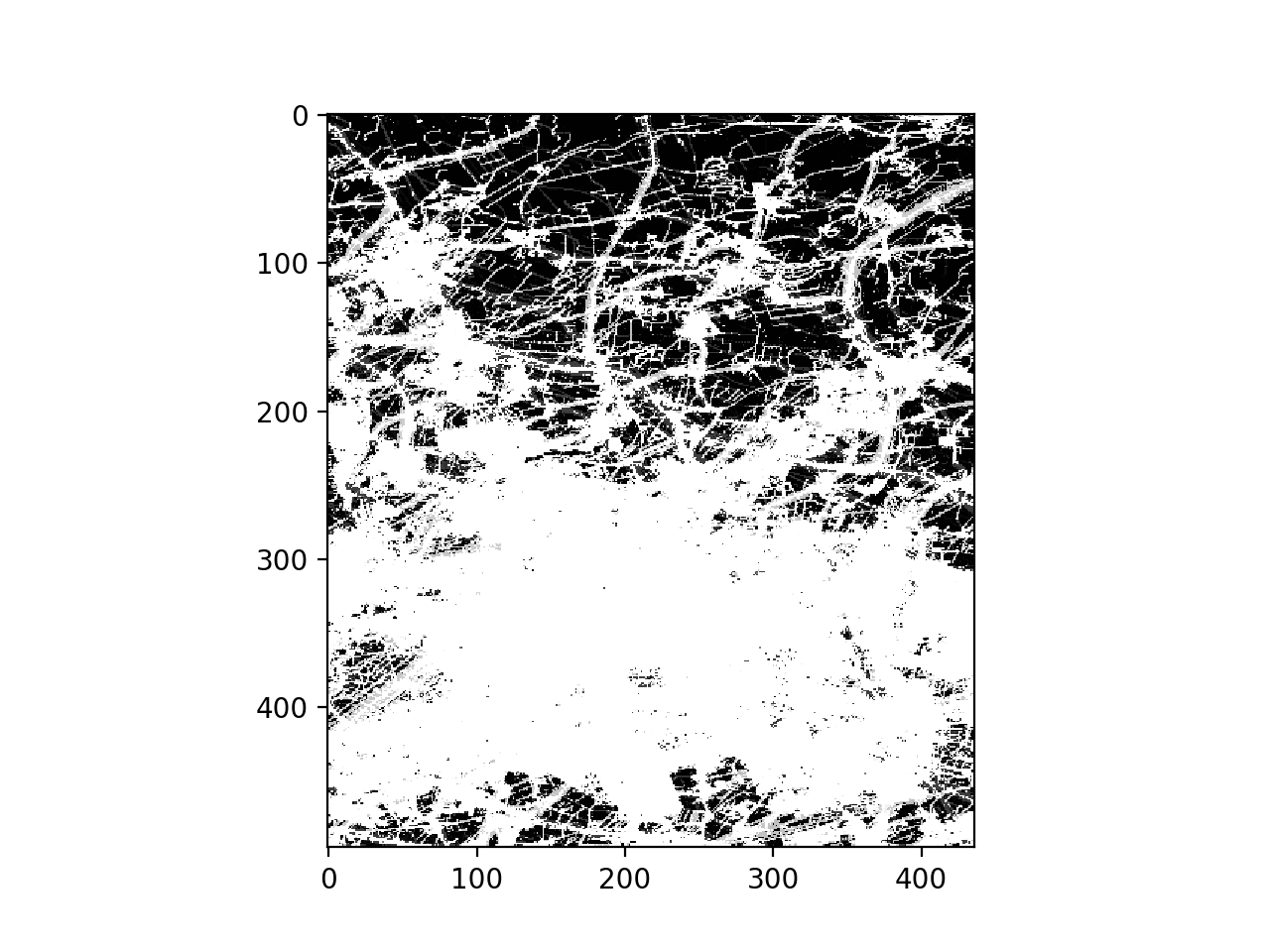}
    \caption{181 days of data}
  \end{subfigure}
  \caption{Berlin binary masks}
  \label{fig:masks}
\end{figure}

% The binary mask was created by setting a pixel to 1 if there had ever been a probe data at that point thus aiming to highlight the roads of interest.
Originally it was assumed that a mask made from a small sample of days would suffice, however it was quickly noticed that the positions of certain roads have a tendency to drift (likely caused by GPS drift). This is best seen from the figures \ref{fig:masks}a and b below highlighting masks for Berlin using one day of data versus the entire training set respectively.
\vspace{-4mm}
\vspace{-7mm}
\section{Results and Discussion}
The following tables show validation MSE scores for the model iterations discussed earlier. The `Final Model' denotes Base Model + Down-sampling with Skip-Connections and `CLR' refers to Cyclical Learning Rate. The Final Test Set MSE score achieved was 0.00127527075484. 
\begin{table}[H]
  \caption{Validation MSE Loss with and without downsampling}
  \label{val_mse}
  \centering
  \begin{tabular}{lll}
    \toprule
    \cmidrule(r){1-2}
    City       & Base Model (BM) &     BM + Down-sampling \\
    \midrule
    Berlin     & \textbf{0.00137628335505}  & 0.00138348212931 \\ 
    Istanbul   & \textbf{0.00101007695775}  & 0.00101161806378 \\
    Moscow     & \textbf{0.00146080681588}  & 0.00146263395436 \\
    \midrule                                  
    Average       & \textbf{0.00128238904289}  & 0.00128591138248\\
    \bottomrule
  \end{tabular}
\end{table}
As seen in Table 2 the base model (autoencoder with a 3 layer encoder and 3 layer decoder), mentioned in section 3.3 Model Architecture, achieved a reasonable MSE with 28 epochs of training using the Adam optimiser and a learning rate of 0.002. Down-sampling was not used initially for fear of a loss in learnt topological features from the process of down-sampling and the subsequent up-sampling. This hypothesis was proven correct, seen by a lower loss shown in Table 2 when down-sampling was introduced. Despite the increase in MSE loss, training time was reduced, owing to the fact that 3 of the 6 Convolutional LSTM layers within the model now had inputs half the size of what the base model originally had. In order to rectify this apparent loss of topological features, skip-connections were introduced in a `U-Net style'. Using the same training approach as before the loss was improved on when compared to the base model. 

In order to achieve further improvement exploring parameter space for an optimal learning rate was required, however given the time commitment required for this approach an alternative method was used. By utilising a cyclical learning rate over 28 epochs oscillating between learning rates of $1\times10^{-7}$ and $2\times10^{-3}$, reductions in MSE loss using the same model architecture waa observed, see Table 3. 

The effectiveness of using cyclical learning rates can be intuitively deduced;
if a saddle point or local minima is encountered, lower learning rates won't provide gradients to escape, however higher learning rates will. Therefore by oscillating between higher and lower learning rates, settling in a local minima or saddle points can be avoided allowing the loss to reach a global minima.
\begin{table}[H]
  \caption{Validation MSE Loss with and without CLR}
  \label{val_mse_clr}
  \centering
  \begin{tabular}{lll}
    \toprule
    \cmidrule(r){1-2}
    City     & Final Model without CLR &     Final Model with CLR \\
    \midrule
    Berlin   & 0.00137459288816  & \textbf{0.00133864406961} \\
    Istanbul & 0.00100649357773  & \textbf{0.00097952096257} \\
    Moscow   & 0.00146005756687  & \textbf{0.00143169029615} \\
    \midrule                      
    Average    &  0.00128038134425  & \textbf{0.00124995177611} \\
    \bottomrule
  \end{tabular}
\end{table}
Given additional time, utilising the available static data may lead to further improvements and reductions to MSE. Using a one-hot encoding with the 9th channel (road incidents) where 0 and 1 represents absence and presence of road incidents respectively could be applied giving insight as to whether traffic remains stationary for a given time after an incident has taken place. An alternative approach would be to scale the road incidents where more incidents results in a higher scaling number, allowing the model to learn responses in a range from minor to extreme cases.  

A similar one-hot encoding approach could also be applied to days of the week, specifying either weekend or weekday, enabling the model to recognise and learn behavioural differences given the type of day.

% The Final Test Set MSE score achieved was: 0.00127527075484 
% add results for build, build skip, build down, build skip down

% talk about progress 

% Base model
% Berlin 3x3 AE : 0.0013834821293130517
% Istanbul 3x3 AE: 0.0010116180637851357
% Moscow 3x3 AE: 0.0014626339543610811

% Base with downsampling
% Berlin no CLR: 0.0013762833550572395
% Istanbul no CLR: 0.0010100769577547908
% Moscow no CLR: 0.0014608068158850074

% with downsampling and skip connections no clr (0.001)
% Berlin base model 2x2: 0.0013745928881689906
% Istanbul base model 2x2: 0.001006493577733636
% Moscow base model 2x2: 0.0014600575668737292

% with down and skip and clr
% 0.001338644069619476
% 0.000979520962573588
% 0.001431690296158194
% Final: 0.00127527075484 
\section{Conclusion and Future Work}
This paper has presented a Temporal Autoencoder with U-Net style skip-connections for the purpose of Traffic Forecasting, presented as a frame prediction task introduced by IARAI's Traffic4Cast 2020 challenge. The model described achieves good performance in predicting average speeds and aggregated volumes at different head-ways across three cities, obtaining a competitive MSE score of 0.00127527075484 on the challenge test set.

In this work, recurrent and computer vision techniques were married together through the utilisation of Convolutional LSTM layers to form the basis for an Autoencoder design with skip-connections similar to U-Net to ensure there was no loss in learnt topological features through down-sampling. A cylical learning rate was also employed to improve training efficiency achieving lower loss scores in fewer epochs than standard approaches of using a set learning rate with a large number of epochs. 

The potential for building practical applications using this methodology could aid in the creation of reliable traffic forecasting systems to ease negative impacts caused by traffic congestion in the ever growing pursuit of sustainable mobility.

Future research should consider the prospects of other generative models and architectures for frame prediction, taking particular notice of adapting existing Image-to-Image translation networks \cite{Isola2017}. Especially those using GAN architectures with U-Net generators and discriminators. Applying recurrent modules such as the Convolutional LSTM to these architectures could prove fruitful.

\clearpage
\bibliographystyle{IEEEtran}
\bibliography{main.bib}

\end{document}